\PassOptionsToPackage{table}{xcolor}
\documentclass[sigconf]{acmart}
\usepackage{algorithm}
\usepackage{algorithmic}

\usepackage{booktabs}
\usepackage{multirow}
\usepackage{pifont}
\usepackage{xspace}
\usepackage{subfigure}
\usepackage{balance}
\newcommand{\ourmethod}{FreeGAD\xspace}

\copyrightyear{2025}
\acmYear{2025}
\setcopyright{acmlicensed}
\acmConference[CIKM '25]{Proceedings of the 34th ACM International Conference on Information and Knowledge Management}{November 10--14, 2025}{Seoul, Republic of Korea}
\acmBooktitle{Proceedings of the 34th ACM International Conference on Information and Knowledge Management (CIKM '25), November 10--14, 2025, Seoul, Republic of Korea}
\acmDOI{10.1145/3746252.3761125}
\acmISBN{979-8-4007-2040-6/2025/11}

\begin{document}
\title{FreeGAD: A Training-Free yet Effective Approach for Graph Anomaly Detection}


\author{Yunfeng Zhao}
\authornote{All authors contributed equally to this research.}
\orcid{0009-0002-5482-4454}
\affiliation{%
\institution{Guangxi University}
\city{Nanning}
\country{China}
}
\email{yunf.zhao@st.gxu.edu.cn}

\author{Yixin Liu}
\authornotemark[1]
\orcid{0000-0002-4309-5076}
\affiliation{%
\institution{Griffith University}
\city{Gold Coast}
\country{Australia}
}
\email{yixin.liu@griffith.edu.au}

\author{Shiyuan Li}
\authornotemark[1]
\orcid{0000-0002-4381-7497}
\affiliation{%
\institution{Guangxi University}
\city{Nanning}
\country{China}
}
\email{li.shiy511@gmail.com}

\author{Qingfeng Chen}
\authornote{Corresponding author.}
\orcid{0000-0002-5506-8913}
\affiliation{%
\institution{Guangxi University}
\city{Nanning}
\country{China}
}
\email{qingfeng@gxu.edu.cn}

\author{Yu Zheng}
\orcid{0000-0003-0757-4210}
\affiliation{%
\institution{Griffith University}
\city{Gold Coast}
\country{Australia}
}
\email{zhengyu511@gmail.com}

\author{Shirui Pan}
\orcid{0000-0003-0794-527X}
\affiliation{%
\institution{Griffith University}
\city{Gold Coast}
\country{Australia}
}
\email{s.pan@griffith.edu.au}

\renewcommand{\shortauthors}{Yunfeng Zhao et al.}


\begin{abstract}
Graph Anomaly Detection (GAD) aims to identify nodes that deviate from the majority within a graph, playing a crucial role in applications such as social networks and e-commerce. 
Despite the current advancements in deep learning-based GAD, existing approaches often suffer from high deployment costs and poor scalability due to their complex and resource-intensive training processes. 
Surprisingly, our empirical findings suggest that the training phase of deep GAD methods, commonly perceived as crucial, may actually contribute less to anomaly detection performance than expected. 
Inspired by this, we propose \textbf{\ourmethod}, a novel training-free yet effective GAD method. Specifically, it leverages an affinity-gated residual encoder to generate anomaly-aware representations. Meanwhile, \ourmethod identifies anchor nodes as pseudo-normal and anomalous guides, followed by calculating anomaly scores through anchor-guided statistical deviations.
Extensive experiments demonstrate that \ourmethod achieves superior anomaly detection performance, efficiency, and scalability on multiple benchmark datasets from diverse domains, without any training or iterative optimization. 
\end{abstract}

\maketitle

\section{Introduction}
\begin{figure}[t]
    \centering
    \subfigure[Runtime comparison]{\label{subfig:intro_1}
        \includegraphics[scale=0.226]{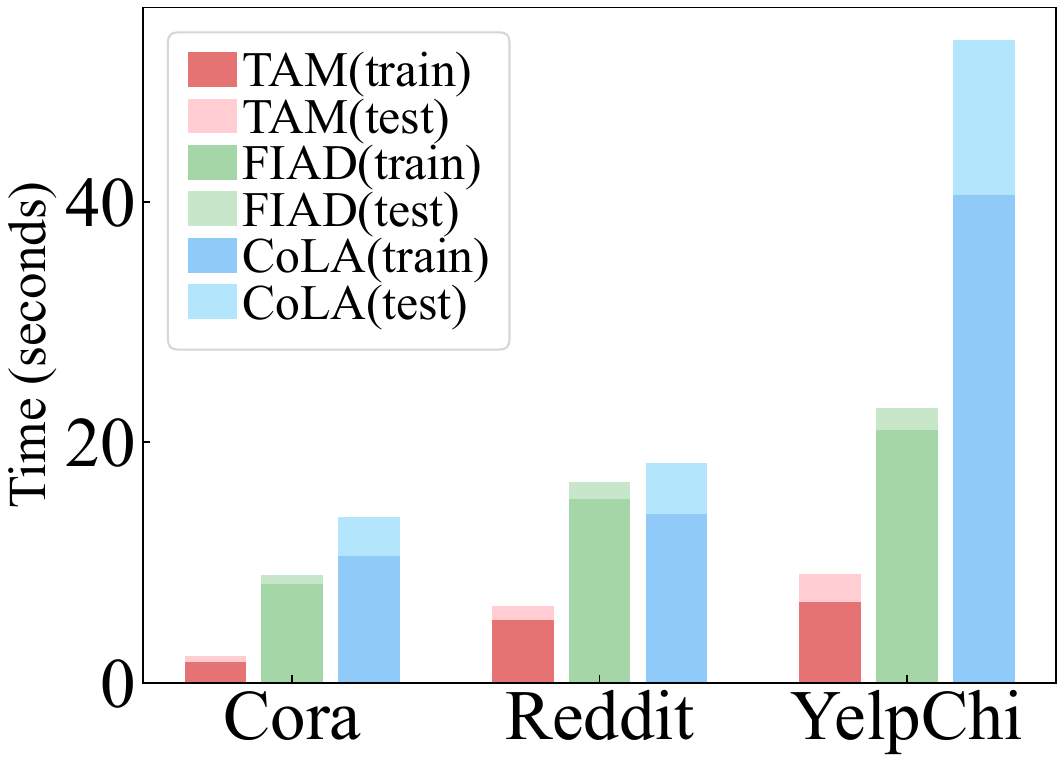}
    }\hfill
    \subfigure[Training-free performance]{\label{subfig:intro_2}
        \includegraphics[scale=0.226]{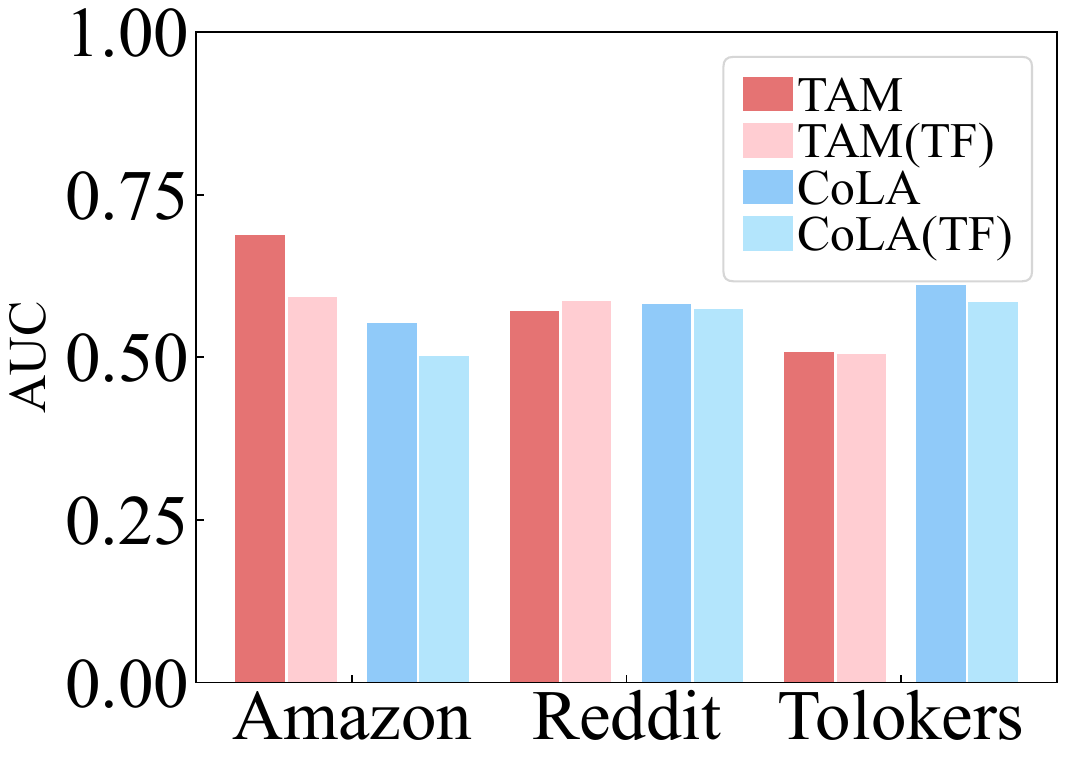}
    }
    \caption{(a): The efficiency of different GAD methods, with the runtime square-rooted for comparison. (b): The difference of GAD methods with and without training.}
    \label{fig:intro}
\end{figure}

Graph anomaly detection (GAD) is a critical research area focused on identifying abnormal nodes that significantly deviate from normal patterns within a graph~\cite{qiao2024deep,ma2021comprehensive}. GAD has been applied to numerous real-world scenarios, such as fraud detection in financial networks, intrusion detection in cybersecurity, and identifying irregularities in social or communication networks~\cite{islam2020graph,pourhabibi2020fraud,yang2021rumor}. The growing importance of GAD in real-world applications has led to increasing research attention being devoted to developing effective and robust methods for detecting anomalies in graph-structured data~\cite{ding2019deep,liu2021anomaly,qiao2023truncated}.

Due to the difficulty of obtaining labeled anomaly data in real-world scenarios, the mainstream GAD methods primarily focus on unsupervised scenarios~\cite{qiao2024deep,qiao2023truncated,chen2025fiad,fan2020anomalydae}. While the early shallow GAD methods often face limitations in handling complex and high-dimensional real-world graph data~\cite{ding2019deep}, deep learning-based GAD methods have emerged in recent years as the de facto solution, demonstrating superior performance in benchmark datasets~\cite{qiao2023truncated,chen2025fiad}. These deep GAD methods leverage graph neural networks (GNNs) as their backbone models and employ anomaly-aware unsupervised techniques or customized objectives for model optimization. After sufficient training, the model output can be transformed into abnormality measurements through specific rules or algorithms. For example, DOMINANT~\cite{ding2019deep} and AnomalyDAE~\cite{fan2020anomalydae} employ graph autoencoder models to reconstruct graph data, using reconstruction errors as indicators of anomalies. CoLA~\cite{liu2021anomaly} and ANEMONE~\cite{jin2021anemone} introduce GNN-based contrastive learning frameworks for GAD, where the degree of mutual agreement between contrastive elements can measure the node abnormality. More recent methods adopt diverse unsupervised learning and anomaly scoring strategies, such as hop counting~\cite{huang2022hop} and local affinity estimation~\cite{qiao2023truncated}, to further raise the bar of GAD performance. 

Despite their effectiveness, the costs associated with training these deep GAD models are non-negligible. To achieve optimal detection performance, hundreds of epochs of training are often necessary, leading to high computational costs and prolonged deployment times. As shown in Fig.~\ref{subfig:intro_1}, the training time constitutes a substantial portion of the deployment duration for most approaches, causing delays in applying anomaly detection to time-sensitive systems and diminishing their practical utility. Moreover, since GNNs require complete graph structures for propagation operations, the training of GNN models typically involves performing backpropagation over the entire graph~\cite{chiang2019cluster}. As a result, training GNN-based GAD models on large-scale datasets becomes highly challenging due to increased memory requirements, further limiting their scalability and practical applicability in real-world scenarios. As evidence, our experiments in Table~\ref{tab:combined} show that most GAD methods fail to run on T-Finance dataset (with 39k nodes) on a GPU with 24GB memory. The aforementioned limitations prompt us to consider a question: \textit{\textbf{Is it truly necessary to heavily train GNN-based GAD models to achieve effective anomaly detection?}}

Recalling the framework architectures of mainstream deep GAD approaches, the learnable components mainly lie in the transformation operations within the GNN model, where features are transformed through learnable projections to refine node representations. Apart from this, other components, such as propagation and anomaly scoring, are typically parameter-free, and thus do not require training for optimization. To verify the contribution of model training to the final anomaly detection performance, we modify two representative GAD methods (i.e., CoLA~\cite{liu2021anomaly} and TAM~\cite{qiao2023truncated}) into training-free (TF) variants by removing the learnable components. We evaluate their performance against the original models and the comparison is demonstrated in Fig.~\ref{subfig:intro_2}. Surprisingly, the training-free variants perform competitively against their fully trained counterparts, with only a 5.0\% drop in average. 
Our observation suggests that model training may not be as critical to anomaly detection performance as previously thought. In this case, a natural question arises: \textbf{\textit{Can we design a GAD approach that achieves competitive anomaly detection performance without the need for training?}}

Motivated by the above question, this paper proposes a training-\textbf{\underline{Free}} yet effective approach for \textbf{\underline{G}}raph \textbf{\underline{A}}nomaly \textbf{\underline{D}}etection, abbreviated as \textbf{\underline{\ourmethod}}. \ourmethod can effectively detect abnormal nodes without the need for training, leveraging a propagation-only encoder to generate node representations and an anchor-based distance measurement module to predict node abnormality. To be more specific, \ourmethod begins with an affinity-gated residual encoder to generate anomaly-aware representations, explicitly capturing the inherent affinity between nodes and their multi-hop neighbors. After that, an anchor node selection module identifies anchor nodes, which are the most representative samples, serving as pseudo normal and anomaly nodes for guiding the model in discriminating between normal and anomalous patterns. Finally, an anchor-guided anomaly scoring module calculates the abnormality score for each node by measuring its statistical deviation from the anchor nodes, enabling precise identification of anomalies. The entire process operates directly on the data in a single run, without requiring any training or optimization. To verify the effectiveness of \ourmethod, we conduct extensive experiments on 10 real-world benchmark datasets, comparing its performance against state-of-the-art GAD methods. The empirical study demonstrates the advantages of \ourmethod: 

\begin{itemize}
    \item \textbf{Superior Detection Performance.} Without the need for training, \ourmethod achieves state-of-the-art performance on 6 out of 10 datasets and competitive performance on the remaining ones. Notably, \ourmethod excels in detecting real anomalies compared to the synthetic ones, which highlights its potential in real-world fraud and intrusion detection scenarios. 
    \item \textbf{Extremely Low Time Cost.} Due to its training-free nature, \ourmethod has significantly shorter deployment time compared to baselines, indicating its applicability for time-sensitive applications.
    \item \textbf{Excellent Scalability.} \ourmethod demonstrates enhanced scalability, as it can be effectively applied to large-scale datasets, such as Elliptic with 200k+ nodes, as well as to the over-dense dataset T-Finance with 21m+ edges, highlighting its capacity to handle complex and large-scale graph data.
\end{itemize}

\section{Related Work}
\subsection{Graph Anomaly Detection}
Graph anomaly detection (GAD) aims to identify the nodes that exhibit anomalous behavior or deviate significantly from the normal patterns in a graph.

Earlier studies primarily focus on using shallow learning approaches for GAD~\cite{akoglu2015graph,li2017radar}. For example, AMEN~\cite{perozzi2016scalable} leveraged the information of ego-network for each node to detect anomalous neighborhoods within attributed networks. ANOMALOUS~\cite{peng2018anomalous} employs a joint framework with CUR
decomposition and residual analysis for GAD. Although these methods do not leverage deep learning techniques, most of them still require iterative optimization for training.

Recent research highlights the effectiveness of deep learning in GAD, utilizing graph neural networks (GNNs) to build powerful GAD models~\cite{zheng2021generative,qiao2024deep}. For instance, DOMINANT~\cite{ding2019deep} employs a graph-convolutional autoencoder (GAE) to reconstruct both the adjacency and attribute matrices, where node anomalies are assessed based on reconstruction errors. SpecAE~\cite{li2019specae} and AnomalyDAE~\cite{fan2020anomalydae} use advanced GAE architectures to boost the performance of DOMINANT. Apart from reconstruction models, another line of studies, such as CoLA~\cite{liu2021anomaly} and ANEMONE~\cite{jin2021anemone}, utilizes contrastive GNN frameworks to learn robust scoring models for anomaly detection. PREM~\cite{pan2023prem} further refines contrastive GAD models by simplifying their architecture and eliminating redundant components. Recently, Qiao et al.~\shortcite{qiao2023truncated} indicate the key role of \textit{local affinity} in detecting anomalous nodes and propose a truncated affinity maximization (TAM) model for GAD. HUGE~\cite{pan2025label} proposes HALO, a label-free heterogeneity measure estimated from node attributes, for unsupervised node anomaly detection. For all deep learning GAD methods, gradient descent-based optimization is required for model training, which can be time-consuming.

While existing deep GAD methods have demonstrated strong performance, they require iterative training or optimization to achieve strong detection performance, leading to increased deployment time and resource demands. To address this problem, the aim of this paper is to propose a learning-free yet effective baseline model for GAD.

\subsection{Graph Neural Networks}
Graph neural networks (GNNs) are a neural network architecture designed for graph-structured data, employing a message-passing mechanism with two key operations: \textit{propagation} (P), which facilitates information exchange between neighboring nodes, and \textit{transformation} (T), which updates node embeddings through learnable non-linear projections~\cite{wu2020comprehensive}. Due to their powerful capabilities, GNNs have been successfully applied across a wide range of tasks~\cite{li2024noise,liu2024arc,liu2024self,shen2025understanding,li2025assemble}. Different GNN models leverage diverse propagations and aggregation functions. Frequently utilized P functions comprise averaging aggregation~\cite{kipf2016semi}, attention aggregation~\cite{velivckovic2017graph}, and LSTM aggregation~\cite{hamilton2017inductive}, while T operations are typically implemented as one or more perceptron layers followed by non-linear activations.

Most GNN-based models~\cite{ding2019deep,liu2021anomaly,chen2025uncertainty,liu2025graph,miao2025blindguard} adopt the PTPT structure, which is constructed by sequentially stacking multiple "P-T" operations. Although this design is widely used, it often faces issues of increased runtime and reduced efficiency, and is also prone to overfitting during the training process. Moreover, in some decoupled GNN models~\cite{zhang2022model}, 
propagation and transformation operations are not simply alternated but instead repeat one operation several times before switching to the other. For example, in models such as APPNP~\cite{gasteiger2018predict}, SGC~\cite{wu2019simplifying}, and AP-GCN~\cite{spinelli2020adaptive}, the propagation and transformation operations can be nested cyclically in multiple levels to enhance the expressive power of the model. This design approach provides a more flexible architecture for GNNs that can adapt to different task requirements.

Building on decoupled models, in this paper, we design an affinity-gated residual encoder with multi-hop propagation while discarding feature transformation, which ensures the generation of high-quality representations without requiring training.

\section{Preliminaries}
\noindent \textbf{Notations.}
A undirected attributed graph can be denoted as $\mathcal{G}=(\mathcal{V},\mathcal{E},\mathbf{X})$, where $\mathcal{V}=\{v_1,\cdots,v_n\}$ represents the set of nodes, $\mathcal{E}$ is the set of edges, and $|\mathcal{V}|=n$, $|\mathcal{E}|=e$, respectively. The features of nodes can be described by the node feature matrix $\mathbf{X} \in \mathbb{R}^{n \times m}$, where $m$ is the dimension of features and the $i$-th row $\mathbf{x}_i$ represents the feature vector of the $i$-{th} node $v_i$. 
The connection between nodes can be represented by an adjacency matrix $\mathbf{A} \in \{0,1\}^{n \times n}$, where the $i,j$-th entry $\mathbf{A}_{ij}=1$ means $v_i$ and $v_j$ are connected and vice versa. 
The symmetric normalization of the adjacency matrix is denoted by $\hat{\mathbf{A}}=\widetilde{\mathbf{D}}^{-\frac{1}{2}} \tilde{\mathbf{A}} \widetilde{\mathbf{D}}^{-\frac{1}{2}}$, where $\tilde{\mathbf{A}}=\mathbf{A}+\mathbf{I}$ represents the adjacency matrix of the undirected graph with the addition of the self-loops and $\widetilde{\mathbf{D}}$ is the diagonal degree matrix of $\tilde{\mathbf{A}}$.

\noindent\textbf{Problem Formulation.} 
Considering the sparsity of anomaly labels, this paper focuses on unsupervised graph node anomaly detection (GAD) scenarios. In unsupervised GAD scenario, the node set $\mathcal{V}$ can be divided into anomalous node set $\mathcal{V}_a$ and normal node set $\mathcal{V}_n$, with $\mathcal{V}=\mathcal{V}_a \cup \mathcal{V}_n$, $\mathcal{V}_a \cap \mathcal{V}_n = \emptyset$, and $|\mathcal{V}_a|=N_a \ll |\mathcal{V}_n|=N_n$. The objective of unsupervised GAD is to learn a scoring function (i.e., GAD model) $\mathcal{F}$ to obtain anomaly score for all nodes in $\mathcal{G}$: $\mathcal{F}(\mathbf{A}, \mathbf{X}) \rightarrow \mathbf{s} \in \mathbb{R}^{n}$, where $\mathbf{s}$ is the anomaly scores for all nodes, with each entry $s_i$ indicating the abnormal degree of node $v_i$, i.e., the probability of $v_i$ belonging to $\mathcal{V}_a$ rather than $\mathcal{V}_n$.

\section{Methodology}
\begin{figure*}
    \centering
        \includegraphics[scale=0.8]{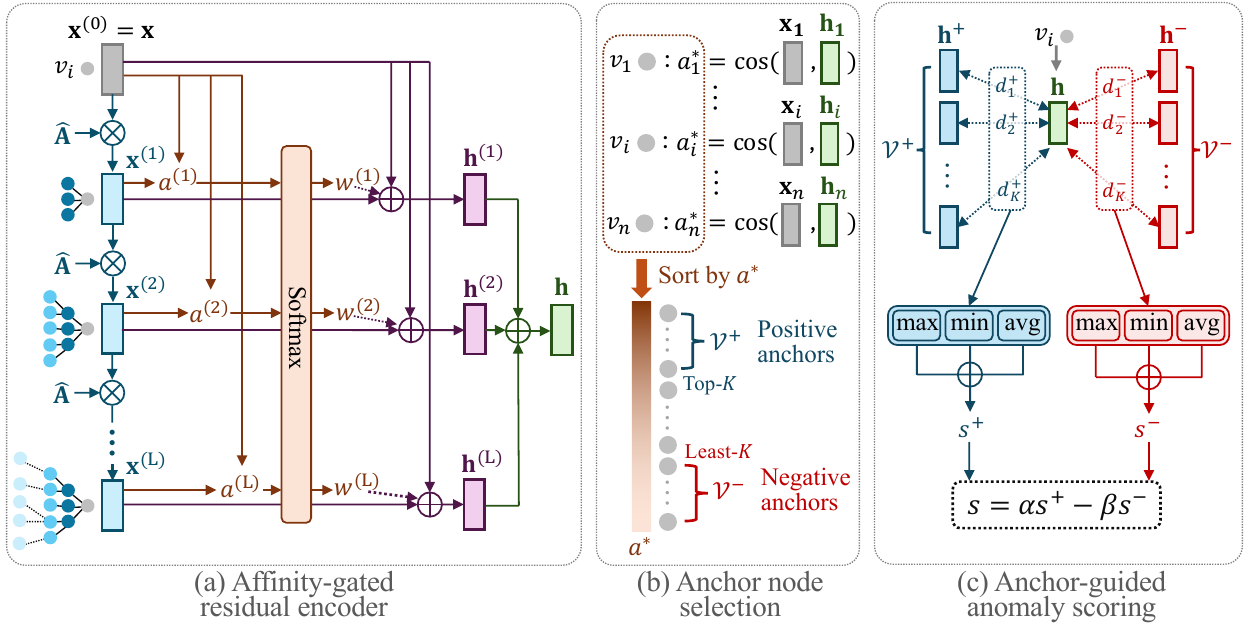}
	\caption{The overall pipeline of \ourmethod. In blocks (a) and (c), we omit the subscript ``${}_i$'' for variables $\mathbf{x}$, $a$, $w$, $\mathbf{h}$, $d$, and $s$ for simplicity.}
	\label{fig:pipeline}
\end{figure*}

In this section, we introduce \ourmethod, a training-free and effective method for graph anomaly detection (GAD). As illustrated in Fig.~\ref{fig:pipeline}, the overall pipeline of \ourmethod consists of three components: \ding{182}~\textit{Affinity-gated residual encoder} to generate anomaly-sensitive node representations~(Sec.~\ref{subsec:encoder}); \ding{183}~\textit{Anchor node selection} to choose the pseudo normal/abnormal nodes that characterize the underlying data distribution~(Sec.~\ref{subsec:select}); and \ding{184}~\textit{Anchor-guided anomaly scoring} to measure node-level abnormality by statistically evaluating the distance between the representations of each node and the selected anchor nodes~(Sec.~\ref{subsec:score}). The following subsections offer a comprehensive explanation of each module.

\subsection{Affinity-Gated Residual Encoder}\label{subsec:encoder}

In the first step of \ourmethod, we design an encoder to generate representations of nodes. The representations are expected to incorporate the semantic and structural information of each node, providing anomaly-related cues for effective abnormality prediction. The state-of-the-art GAD methods (e.g., CoLA~\cite{liu2021anomaly} and TAM~\cite{qiao2023truncated}) usually employ GCN~\cite{kipf2016semi} as encoder, which may suffer from two limitations. \textit{Firstly}, stacking many layers can degrade the representation capability of GCN due to over-smoothing~\cite{chen2020measuring} and model degradation issues~\cite{zhang2022model}.  This limitation hinders GCN from capturing high-order information from long-range nodes, which is crucial for detecting real-world anomalies. \textit{Secondly}, GCN learns representations by simply averaging ego and neighbor information, which may fail to capture anomaly-sensitive knowledge (e.g.  affinity~\cite{qiao2023truncated}). 

To generate high-quality and task-specific representations, we design an advanced, training-free encoder. To address the first limitation, we use a multi-hop propagation-based architecture without feature transformation, which prevents model degradation; also, we introduce a residual design to alleviate over-smoothing problem. To address the second one, we explicitly estimate the affinity from the propagated features and use the affinity to gate the residual connection. In this way, the output affinity-aware representations can better indicate the node-level abnormality. The following paragraphs introduce the detailed procedure of each step.

\noindent\textbf{Multi-Hop Propagation.} 
To capture high-order information in a training-free manner, in \ourmethod, we focus exclusively on propagating node feature representations without introducing additional feature transformation operation with learnable parameters. Specifically, we can achieve this by repeatedly performing message propagation operations:

\begin{equation}
\mathbf{X}^{(l)}=\hat{\mathbf{A}} \mathbf{X}^{(l-1)},
\label{eq:propagation}
\end{equation}

\noindent where $l \in \{1,\ldots,L\}$ is the layer index, $L$ is a pre-defined number of layers, $\mathbf{X}^{(l)}$ is the propagated feature matrix at the $l$-th layer, and $\hat{\mathbf{A}}$ is the normalized adjacency matrix. The features at the first input layer are denoted as the original features, i.e., $\mathbf{X}^{(0)} = \mathbf{X}$. 
The propagation-only design not only prevents the model degradation issues caused by excessive transformations~\cite{zhang2022model}, but also achieves our training-free design goal by removing the transformations.

\noindent\textbf{Affinity Estimation.} 
As highlighted by Qiao et al.~\shortcite{qiao2023truncated}, node affinity is a key property that strongly correlates with node abnormality. While they only focus on local affinity between each node and its 1-hop neighbors, in real-world data, the abnormality can also related to high-order affinity, i.e., the similarity between each node and its high-order neighbors~\cite{liu2024arc}. To capture such patterns in the learned representations, in \ourmethod, we estimate multi-order affinity and leverage it during representation generation. 

Specifically, we quantify neighborhood-specific affinity by measuring the similarity between the raw features and the propagated features derived from different neighborhoods. For node $v_i$, the similarity between the $l$-th propagated feature and the raw feature is calculated as follows:
\begin{equation}\label{eq:sim}
{a}^{(l)}_{i} = \frac{\sum_{j=1}^{m}{\mathbf{x}^{(0)}_{i j} \cdot \mathbf{x}^{(l)}_{i j}}}{{||\mathbf{x}^{(0)}_{i}||}^{2} \cdot {||\mathbf{x}^{(l)}_{i}||}^{2} + \sigma},
\end{equation}
where $|| \cdot ||$ represents the $l2$ norm, $\mathbf{x}^{(l)}_{ij}$ stands for the $i$-th node's $j$-th feature, and the $\sigma$ represents a small value to prevent denominator tend to 0. After obtaining the feature similarity between a node and its multi-hop neighborhoods, we conduct a layer-wise Softmax-based normalization to obtain the affinity weights:

\begin{equation}\label{eq:weight}
{w}^{(l)}_{i} = \frac{\exp\left({a}^{(l)}_{i}\right)}{\sum_{k=1}^{L} \exp\left({a}^{(k)}_{i}\right)}.
\end{equation}

\noindent\textbf{Affinity-Gated Residual.} 
Once we obtain the normalized affinity at each propagation step, we employ the affinity as a gating scalar for the residual operation between propagated features and original features. Specifically, the representation at the $l$-th layer, denoting as $\mathbf{h}_{i}^{(l)}$, can be written as:

\begin{equation}\label{eq:res}
\mathbf{h}_{i}^{(l)} = (1-w_{i}^{(l)}) \mathbf{x}_{i}^{(l)} + w_{i}^{(l)} \mathbf{x}_{i}^{(0)}.
\end{equation}

According to Eq.~\eqref{eq:res}, the representation of each node can be expressed as the linear combination of its propagated and original features. The gating mechanism ensures that, if a node has a higher affinity with its $l$-hop neighbors, then its $l$-layer representation will be closer to the original feature, preserving its original attributes; conversely, if a node exhibits lower affinity with its $l$-hop neighbors, its $l$-layer representation will incorporate more information from the propagated features, emphasizing the differences. Such differential treatment not only effectively distinguishes nodes with high and low affinity (which reflects the degree of node abnormality) at the representation level, but also preserves the ego information within the features. Moreover, the residual operation helps mitigate the over-smoothing problem caused by multi-hop propagation, ensuring the informativeness of the representations.

\noindent\textbf{Multi-Hop Mixing.}
To unify the representations learned at different propagation steps into a single representation, we perform multi-hop mixing by averaging the representations from each layer:
\begin{equation}\label{eq:res_mean}
\mathbf{h}_{i} =  \frac{1}{L}\sum_{l=1}^{L} \mathbf{h}_{i}^{(l)},
\end{equation}

\noindent where $\mathbf{h}_{i}$ is the final representation vector for node $v_i$. With the mixing operations, the information from different propagation steps is effectively aggregated, enabling the final representation to capture both local and global contextual information. 

The representations generated by our affinity-gated residual encoder enjoy several notable merits. \ding{182}~\textbf{Multi-scale-aware}: The multi-hop mixing mechanism incorporates information from multi-hop neighbors, which enriches the representation with broader contextual insights. \ding{183}~\textbf{Affinity-aware}: Thanks to the affinity-gated residual mechanism, the learned representations effectively capture each node's affinity with its multi-hop neighbors, thereby reflecting its abnormality accurately. \ding{184}~\textbf{Training-free}: Our encoder has no learnable parameters, eliminating the need for training in the entire learning process. \ding{185}~\textbf{Original space-preserving}: All operations are performed in the original feature space, which facilitates the computation of affinity for anchor node selection in the next subsection.

\subsection{Anchor Node Selection}\label{subsec:select}
After obtaining the node representations, the next key step is to predict the abnormality (i.e., anomaly score) of each node. However, in unsupervised GAD settings, the inaccessibility of labels complicates the direct supervision of the scoring module. As a result, it is necessary to design a training-free mechanism that can effectively estimate the anomaly scores without relying on labeled data for \ourmethod.

According to the idea of pseudo-label self-supervised learning~\cite{sun2020multi}, we design an anchor node-guided anomaly scoring algorithm. Our rationale is to filter out the most representative nodes from the graph as \textbf{\textit{anchor nodes}}, and then measure the degree of abnormality of each node based on its distance to the anchor nodes. To be more specific, we incorporate two types of anchor nodes: \textbf{{positive anchor nodes}} corresponding to the pseudo-normal nodes in the graph, and \textbf{negative anchor nodes} corresponding to the nodes that exhibit abnormal characteristics. To select the anchor nodes with a simple criterion, we again rely on the affinity measurement due to its strong correlation with abnormality. Based on the similarity between the representation and raw feature of each node $v_i$, its affinity $a_{i}^{*}$ can be calculated by:

\begin{equation}\label{eq:a_star}
a_{i}^{*} = \operatorname{sim}(\mathbf{x}_{i}, \mathbf{h}_{i}),
\end{equation}

\noindent where $\operatorname{sim}(\cdot)$ represents the similarity function defined according to Eq.~\eqref{eq:sim}. By collecting the affinity of all nodes, we can obtain an affinity list, denoted by $\mathbf{a}^{*}=[a_{1}^{*}, a_{2}^{*}, \ldots, a_{n}^{*}]$. Then, we rearrange the list in descending order based on the affinity values, which can be denoted as $\mathbf{a}^{*}_{sorted}=\operatorname{sort}(\mathbf{a}^{*})$. According to the sorted list, we select the top-$K$ nodes with the highest affinity values as the positive anchor nodes:

\begin{equation}\label{eq:anchor_pos}
\mathcal{V}^{+} = \{v_i \in \mathcal{V} | \operatorname{idx}(a_{i}^{*}, \mathbf{a}^{*}_{sorted}) \le {K}\},
\end{equation}

\noindent where $\mathcal{V}^{+}=\{v_{1}^{+},v_{2}^{+},\ldots,v_{K}^{+}\}$ is the set of positive anchor nodes and $\operatorname{idx}(a_{i}^{*},\mathbf{a}^{*}_{sorted})$ returns the index of $a_{i}^{*}$ within $\mathbf{a}^{*}_{sorted}$. In the same way, the negative anchor nodes can be selected as the top-${K}$ nodes with the smallest affinity values: 

\begin{equation}\label{eq:anchor_neg}
\mathcal{V}^{-} = \{v_i \in \mathcal{V} | \operatorname{idx}(a_{i}^{*}, \mathbf{a}^{*}_{sorted}) \ge n-{K}+1\},
\end{equation}

\noindent where $\mathcal{V}^{-}=\{v_{1}^{-},v_{2}^{-},\ldots,v_{K}^{-}\}$ denotes the set of negative anchor nodes. Through the simple selection algorithm, the most representative nodes in a graph can be identified and then serve as a critical clue to predict the abnormality of each node.

\subsection{Anchor-Guided Anomaly Scoring}\label{subsec:score}

After collecting positive anchor nodes (i.e. pseudo-normal nodes) and negative anchor nodes (i.e. pseudo-abnormal nodes), we utilize an anchor-guided anomaly scoring algorithm to generate the anomaly scores via distance measurement. Our basic assumption is that normal nodes share common patterns with each other and, hence, have closer distances in the representation space; similarly, anomalies may have different shared patterns, leading to their closer representation distances. Therefore, a node that is closer to the positive anchor nodes is potentially more likely to be normal, and vice versa. Motivated by this, we propose to leverage the statistic of the distance measurement between a node $v_i$ and the anchor nodes $\mathcal{V}^+$/$\mathcal{V}^-$ to estimate the anomaly score of $v_i$.

Formally, given a node $v_i$ and a positive anchor node $v^+_k$, we can calculate the Euclidean distance $d_{ik}^+$ between their representations, which can be denoted by:

\begin{equation}\label{eq:dist}
d_{ik}^+ = \| \mathbf{h}_i - \mathbf{h}_k^+ \|_2.
\end{equation}

\noindent In the same way, we can obtain $v_i$'s distance to all positive and negative anchor nodes, denoted as $\mathcal{D}_{i}^{+} = \{d_{i 1}^{+}, \cdots, d_{i {K}}^{+}\}$ and $\mathcal{D}_{i}^{-} = \{d_{i 1}^{-}, \cdots, d_{i {K}}^{-}\}$, respectively. Considering the diversity of anchor nodes, we combine multiple statistics of the distances $\mathcal{D}_{i}^{+}$ and $\mathcal{D}_{i}^{-}$ to obtain a more comprehensive anomaly assessment. Concretely, the positive score ${s}_{i}^{+}$ for node $v_i$ can be computed by:

\begin{equation}\label{eq:score_pos}
    s_{i}^{+}= \operatorname{min}(\mathcal{D}_{i}^{+}) + \operatorname{max}(\mathcal{D}_{i}^{+}) + \operatorname{avg}(\mathcal{D}_{i}^{+}),
\end{equation}

\noindent where $\operatorname{min}(\cdot)$, $\operatorname{max}(\cdot)$, and $\operatorname{avg}(\cdot)$ extract the minimum, maximum, and average values of the positive distance set $\mathcal{D}_{i}^{+}$, respectively. In a similar way, the negative score ${s}_{i}^{-}$ can be calculated according to the negative distance set $\mathcal{D}_{i}^{-}$. Finally, we perform a weighted sum of the positive and negative scores to obtain the final anomaly score of the node $v_i$:
\begin{equation}\label{eq:score}
    s_{i}= \alpha s_{i}^{+}-  \beta s_{i}^{-} ,
\end{equation}

\noindent where $\alpha$ and $\beta$ are hyper-parameters between 0 and 1. 

It is worth noting that there are no learnable parameters involved in the entire architecture of \ourmethod, which means that the model does not require any training and learning. It also avoids overfitting to specific anomaly patterns caused by training towards specific data. Compared with existing GAD methods such as those based on data reconstruction and the comparative GAD methods that include negative score computation and multiple rounds of estimation~\cite{jin2021anemone,liu2021anomaly}, \ourmethod also provides higher running efficiency in the anomaly score estimation process. 

In general, the overall procedures of \ourmethod are shown in Algo.~\ref{algo_1}.

\begin{algorithm}[tb]
  \caption{The Overall Procedure of \ourmethod}
  \label{algo_1}
  \begin{algorithmic}[1]
    \REQUIRE Graph $\mathcal{G}$, Propagation iteration: $L$, Number of anchor nodes: $K$, Positive score weight: $\alpha$, Negative score weight: $\beta$.
    \ENSURE Anomaly score ${s}$.
    \STATE Extract feature matrix $\mathbf{X}$, adjacency matrix $\mathbf{A}$ from $\mathcal{G}$.
    \STATE \textit{// Affinity-gated residual encoder.}
    \FOR{$l = 1\text{:}L$}
      \STATE $\mathbf{X}^{(l)} \gets \text{Propagation}(\mathbf{X}, \mathbf{A}; L)$ via Eq.~\eqref{eq:propagation}
    \ENDFOR
    \FOR{$\mathbf{x}_{i} \in \mathbf{X}$}
      \FOR{$l = 1\text{:}L$}
        \STATE $a_{i}^{(l)} \gets$ Similarity($\mathbf{x}^{(0)}_{i}, \mathbf{x}^{(l)}_{i}$) via Eq.~\eqref{eq:sim}
      \ENDFOR
      \STATE Calculate $W_{i} = \{w_{i}^{(1)}, \dots, w_{i}^{(l)}\} \gets \mathrm{Softmax}(\mathbf{a}_{i})$ via Eq.~\eqref{eq:weight}
      \FOR{$l = 1\text{:}L$}
        \STATE $\mathbf{h}_{i}^{(l)} \gets$ Gating mechanism via Eq.~\eqref{eq:res}
      \ENDFOR
      \STATE $\mathbf{h}_{i} \gets \mathrm{Mean}(\mathbf{h}_{i}^{(1)}, \dots, \mathbf{h}_{i}^{(l)})$ via Eq.~\eqref{eq:res_mean}
    \ENDFOR
    \STATE \textit{// Anchor node selection.}
    \STATE Calculate the affinity $\mathbf{a}^{*}=[a_{1}^{*}, a_{2}^{*}, \ldots, a_{n}^{*}]$ of raw feature $[\mathbf{x}_{1}, \mathbf{x}_{2}, \ldots, \mathbf{x}_{n}]$ and representation $[\mathbf{h}_{1}, \mathbf{h}_{2}, \ldots, \mathbf{h}_{n}]$ via Eq.~\eqref{eq:a_star}
    
    \STATE The set of positive anchor nodes $\mathcal{V}^{+}=\{v_{1}^{+},v_{2}^{+},\ldots,v_{K}^{+}\} \gets \text{Top}(a^*, K, \mathbf{h})$ via Eq.~\eqref{eq:anchor_pos}
    \STATE The set of negative anchor nodes $\mathcal{V}^{-}=\{v_{1}^{-},v_{2}^{-},\ldots,v_{K}^{-}\} \gets \text{Bottom}(a^*, K, \mathbf{h})$ via Eq.~\eqref{eq:anchor_neg}
    \STATE \textit{// Anchor-guided anomaly scoring.}
    \FOR{$\mathbf{h}_{i} \in \mathbf{h}$}
      \STATE $\mathcal{D}_{i}^{+} = \{d_{i1}^+, \dots, d_{ik}^+\}$ as the distances between the target node $\mathbf{h}_{i}$ and each of the $K$ positive anchor nodes.
      \STATE $\mathcal{D}_{i}^{-} = \{d_{i1}^-, \dots, d_{ik}^-\}$ as the distances between the target node $\mathbf{h}_{i}$ and each of the $K$ negative anchor nodes.
      \STATE Calculate the predicted scores $(s_{i}^{+}, s_{i}^{-})$ of anchor node via Eq.~\eqref{eq:score_pos}
      \STATE Calculate the anomaly score ${s}_{i}$ via Eq.~\eqref{eq:score}
    \ENDFOR
  \end{algorithmic}
\end{algorithm}
\subsection{Complexity Analysis}\label{subsec:complex}
The time complexity of \ourmethod primarily consists of three main components: \ding{182} \textbf{Affinity-gated residual encoder.} The overall time complexity for this component is $\mathcal{O}(emL + nL + 3nmL)$, where $e$, $n$, and $m$ represent the number of edges, nodes and the dimension of the feature, respectively, and $L$ is the number of propagation layers. The term $emL$ accounts for the feature propagation process, while the term $nmL$ is the complexity associated with feature similarity calculation, weighted summation, and the average of different propagation layers. 
\ding{183} \textbf{Anchor node selection.} The time complexity for the selection of anchor nodes is $\mathcal{O}(nm + 2nK)$, where $K$ denotes the number of selected anchor nodes (positive and negative). The first term $nm$, corresponds to the calculation of the similarity between the node representation and its raw feature, while the second part, $2nK$, represents the process of selecting the anchor nodes. 
\ding{184} \textbf{Anchor-guided anomaly scoring.} Finally, the time complexity to calculate the anomaly score using the anchor-guided approach is $\mathcal{O}(nmK)$. Therefore, the overall time complexity of \ourmethod is $\mathcal{O}(m(eL+3nL+n+nK)+nL+2nK)$.

\section{Experiments}
\begin{table}
\caption{Statistics of the datasets.}
\centering
\begin{tabular}{l r r r r} 
\toprule
\textbf{Name} & \textbf{\#Nodes} & \textbf{\#Edges} & \textbf{\#Dim.} & \textbf{\%Ano.} \\ 
\midrule
\multicolumn{5}{c}{\cellcolor[rgb]{0.851,0.851,0.851} \textbf{Dataset with real anomalies}} \\ 
Amazon & 10,244 & 175,608 & 25 & 6.76 \\ 
Reddit & 10,984 & 168,016 & 64 & 3.33 \\ 
Tolokers & 11,758 & 519,000 & 10 & 21.8 \\ 
YelpChi & 23,831 & 49,315 & 32 & 5.10 \\ 
T-Finance & 39,357 & 21,222,543 & 10 & 4.60 \\ 
Questions & 48,921 & 153,540 & 301 & 2.98 \\ 
Elliptic & 203,769 & 234,355 & 166 & 2.23 \\ 
\midrule
\multicolumn{5}{c}{\cellcolor[rgb]{0.851,0.851,0.851} \textbf{Datasets with injected anomalies}} \\ 
Cora & 2,708 & 5,429 & 1,433 & 5.53 \\ 
BlogCatalog & 5,196 & 171,743 & 8,189 & 5.77 \\
Flickr & 7,575 & 239,738 & 12,047 & 5.94 \\ 
\bottomrule
\end{tabular}
\label{tab:dsets}
\end{table}

\begin{table*}[ht]
\caption{Anomaly detection performance in terms of AUROC and AUPRC (in percent, mean$\pm$std). The best results are highlighted in bold and underlined. OOM:~out-of-memory (24GB GPU). OOT:~out-of-time (12 hours). 
We do not report the standard deviation of \ourmethod since it is a training-free method, and its results are deterministic, without any randomness introduced by training processes or initialization.}
  \centering
  \resizebox{\textwidth}{!}{
\begin{tabular}{c|c|cccccccccc}
  \toprule
\textbf{Metric} &
\textbf{Method} & \textbf{Amazon} & \textbf{Reddit} & \textbf{Tolokers} & \textbf{YelpChi} & \textbf{T-Finance} & \textbf{Questions} & \textbf{Elliptic} & \textbf{Cora} & \textbf{BlogCatalog} & \textbf{Flickr} \\
      \midrule
      \multirow{11}{*}{\textbf{AUROC}}
        & Radar      & 55.43$\scriptstyle\pm0.00$ & 51.22$\scriptstyle\pm0.00$ & 51.71$\scriptstyle\pm0.00$ & 50.61$\scriptstyle\pm0.00$ & OOM & OOM & OOM & 69.32$\scriptstyle\pm0.00$ & 61.16$\scriptstyle\pm0.00$ & 59.52$\scriptstyle\pm0.00$ \\
        & ANOMALOUS  & 55.46$\scriptstyle\pm0.00$ & 55.95$\scriptstyle\pm4.78$ & 46.48$\scriptstyle\pm0.28$ & 48.77$\scriptstyle\pm1.04$ & OOM & OOM & OOM & 69.38$\scriptstyle\pm0.00$ & 70.86$\scriptstyle\pm0.01$ & 59.52$\scriptstyle\pm0.00$ \\
        \cmidrule{2-12}
        & DOMINANT   & 49.56$\scriptstyle\pm0.65$ & 55.70$\scriptstyle\pm0.77$ & 54.43$\scriptstyle\pm0.11$ & 41.87$\scriptstyle\pm1.87$ & OOM & OOM & OOM & 88.92$\scriptstyle\pm0.14$ & 76.34$\scriptstyle\pm0.29$ & 77.19$\scriptstyle\pm1.59$ \\
        & AnomalyDAE & 51.60$\scriptstyle\pm3.31$ & 55.18$\scriptstyle\pm0.43$ & 52.20$\scriptstyle\pm2.94$ & 59.45$\scriptstyle\pm2.32$ & OOT & OOT & OOT & 80.66$\scriptstyle\pm0.25$ & 75.61$\scriptstyle\pm0.02$ & 74.87$\scriptstyle\pm0.09$ \\
        & CoLA       & 55.23$\scriptstyle\pm0.94$ & \textbf{\underline{58.24}}$\scriptstyle\pm0.44$ & 61.17$\scriptstyle\pm0.37$ & 36.46$\scriptstyle\pm0.58$ & 23.45$\scriptstyle\pm0.19$ & 46.37$\scriptstyle\pm0.37$ & OOM & 89.59$\scriptstyle\pm0.31$ & 78.30$\scriptstyle\pm0.21$ & 70.37$\scriptstyle\pm0.39$ \\
        & CONAD      & 51.02$\scriptstyle\pm0.23$ & 54.11$\scriptstyle\pm0.59$ & 49.47$\scriptstyle\pm2.36$ & 56.65$\scriptstyle\pm1.07$ & OOM & OOM & OOM & 71.82$\scriptstyle\pm0.02$ & 72.54$\scriptstyle\pm0.02$ & 72.82$\scriptstyle\pm0.02$ \\
        & TAM        & 68.87$\scriptstyle\pm0.72$ & 57.17$\scriptstyle\pm0.13$ & 50.88$\scriptstyle\pm0.00$ & OOM & OOM & OOM & OOM & 92.35$\scriptstyle\pm0.23$ & 75.58$\scriptstyle\pm0.20$ & 73.26$\scriptstyle\pm0.45$ \\
        & PREM       & 62.61$\scriptstyle\pm11.85$ & 51.87$\scriptstyle\pm6.15$ & 53.48$\scriptstyle\pm4.04$ & 73.93$\scriptstyle\pm1.51$ & 75.30$\scriptstyle\pm12.31$ & 57.16$\scriptstyle\pm1.29$ & 71.23$\scriptstyle\pm11.08$ & \textbf{\underline{95.11}}$\scriptstyle\pm0.23$ & 75.06$\scriptstyle\pm0.74$ & \textbf{\underline{86.53}}$\scriptstyle\pm0.37$ \\
        & GADAM      & 67.09$\scriptstyle\pm3.12$ & 58.18$\scriptstyle\pm0.38$ & 55.42$\scriptstyle\pm0.56$ & 65.53$\scriptstyle\pm1.66$ & 80.20$\scriptstyle\pm7.85$ & 43.70$\scriptstyle\pm0.37$ & 56.42$\scriptstyle\pm0.35$ & 93.93$\scriptstyle\pm0.25$ & \textbf{\underline{81.58}}$\scriptstyle\pm0.18$ & 72.26$\scriptstyle\pm0.21$ \\
        & FIAD       & 52.36$\scriptstyle\pm8.07$ & 55.72$\scriptstyle\pm1.03$ & 49.73$\scriptstyle\pm1.98$ & 43.27$\scriptstyle\pm3.19$ & OOM & OOM & OOM & 87.43$\scriptstyle\pm1.31$ & 75.47$\scriptstyle\pm2.02$ & 77.01$\scriptstyle\pm2.01$ \\
        \cmidrule{2-12}
        & \ourmethod & \textbf{\underline{88.57}} & 57.21 & \textbf{\underline{67.35}} & \textbf{\underline{78.55}} & \textbf{\underline{92.13}} & \textbf{\underline{64.27}} & \textbf{\underline{77.25}} & 84.85 & 74.84 & 74.73 \\
      \midrule
      \multirow{11}{*}{\textbf{AUPRC}}
        & Radar      &  7.30$\scriptstyle\pm0.00$ &  3.53$\scriptstyle\pm0.00$ & 23.24$\scriptstyle\pm0.00$ &  5.88$\scriptstyle\pm0.00$ & OOM & OOM & OOM & 11.86$\scriptstyle\pm0.00$ & 14.80$\scriptstyle\pm0.00$ & 19.88$\scriptstyle\pm0.00$ \\
        & ANOMALOUS  &  7.30$\scriptstyle\pm0.00$ &  3.86$\scriptstyle\pm0.56$ & 20.14$\scriptstyle\pm0.00$ &  5.51$\scriptstyle\pm0.15$ & OOM & OOM & OOM & 25.23$\scriptstyle\pm0.00$ & 30.62$\scriptstyle\pm0.00$ & 19.88$\scriptstyle\pm0.00$ \\
        \cmidrule{2-12}
        & DOMINANT   &  6.08$\scriptstyle\pm0.08$ &  3.71$\scriptstyle\pm0.06$ & 27.95$\scriptstyle\pm0.03$ &  4.24$\scriptstyle\pm0.34$ & OOM & OOM & OOM & 37.81$\scriptstyle\pm0.44$ & 34.55$\scriptstyle\pm0.05$ & 37.10$\scriptstyle\pm0.05$ \\
        & AnomalyDAE &  7.30$\scriptstyle\pm1.18$ &  3.71$\scriptstyle\pm0.02$ & 26.82$\scriptstyle\pm1.12$ &  7.53$\scriptstyle\pm0.59$ & OOT & OOT & OOT & 26.71$\scriptstyle\pm0.61$ & 23.01$\scriptstyle\pm0.09$ & 16.78$\scriptstyle\pm0.24$ \\
        & CoLA       &  8.28$\scriptstyle\pm0.43$ &  4.01$\scriptstyle\pm0.05$ & 27.25$\scriptstyle\pm0.31$ &  3.75$\scriptstyle\pm0.06$ &  2.79$\scriptstyle\pm0.05$ &  2.55$\scriptstyle\pm0.02$ & OOM & 45.79$\scriptstyle\pm2.66$ & 27.09$\scriptstyle\pm0.98$ & 21.36$\scriptstyle\pm0.31$ \\
        & TAM        & 29.73$\scriptstyle\pm1.28$ &  4.32$\scriptstyle\pm0.03$ & 22.98$\scriptstyle\pm0.00$ & OOM & OOM & OOM & OOM & 48.24$\scriptstyle\pm1.54$ & 33.88$\scriptstyle\pm0.77$ & 24.70$\scriptstyle\pm1.87$ \\
        & PREM       &  8.96$\scriptstyle\pm2.38$ &  3.95$\scriptstyle\pm0.73$ & 23.25$\scriptstyle\pm1.67$ & 10.67$\scriptstyle\pm1.34$ & 33.05$\scriptstyle\pm24.39$ &  3.49$\scriptstyle\pm0.24$ &  \textbf{\underline{7.33}}$\scriptstyle\pm3.38$ & 65.99$\scriptstyle\pm0.23$ & 37.31$\scriptstyle\pm0.74$ & \textbf{\underline{46.37}}$\scriptstyle\pm0.37$ \\
        & GADAM      & 15.87$\scriptstyle\pm7.48$ &  \textbf{\underline{5.01}}$\scriptstyle\pm0.23$ & 23.65$\scriptstyle\pm0.35$ & 10.14$\scriptstyle\pm0.32$ & 15.72$\scriptstyle\pm6.12$ &  2.52$\scriptstyle\pm0.10$ &  3.03$\scriptstyle\pm0.05$ & \textbf{\underline{71.83}}$\scriptstyle\pm1.91$ & \textbf{\underline{37.36}}$\scriptstyle\pm0.36$ & 23.19$\scriptstyle\pm0.43$ \\
        & CONAD      &  6.23$\scriptstyle\pm0.03$ &  4.31$\scriptstyle\pm0.10$ & 21.13$\scriptstyle\pm0.11$ &  7.19$\scriptstyle\pm0.44$ & OOM & OOM & OOM & 24.86$\scriptstyle\pm0.05$ & 33.01$\scriptstyle\pm0.00$ & 37.57$\scriptstyle\pm0.01$ \\
        & FIAD       &  6.71$\scriptstyle\pm1.23$ &  3.74$\scriptstyle\pm0.04$ & 29.31$\scriptstyle\pm0.82$ &  4.32$\scriptstyle\pm0.35$ & OOM & OOM & OOM & 29.31$\scriptstyle\pm3.69$ & 25.44$\scriptstyle\pm6.53$ & 19.55$\scriptstyle\pm3.56$ \\
        \cmidrule{2-12}
        & \ourmethod 
        & \textbf{\underline{75.06}}
        & 3.85 
        & \textbf{\underline{32.17}}
        & \textbf{\underline{15.80}} 
        & \textbf{\underline{73.71}}
        & \textbf{\underline{7.01}}
        & 6.51 
        & 49.75 
        & 34.03 
        & 38.68 \\
\bottomrule
    \end{tabular}
  }
  \label{tab:combined}
\end{table*}
\begin{table*}[ht]
\caption{Training and testing time of baselines and \ourmethod in seconds. The shortest overall times are highlighted in bold and underlined.}
\centering
\resizebox{\textwidth}{!}{%
\begin{tabular}{l|cc|cc|cc|cc|cc|cc|cc}
\toprule
\multirow{2}{*}{\textbf{Method}} & \multicolumn{2}{c|}{\textbf{Amazon}} & \multicolumn{2}{c|}{\textbf{Reddit}} & \multicolumn{2}{c|}{\textbf{Tolokers}} & \multicolumn{2}{c|}{\textbf{YelpChi}} & \multicolumn{2}{c|}{\textbf{Cora}} & \multicolumn{2}{c|}{\textbf{BlogCatalog}} & \multicolumn{2}{c}{\textbf{Flickr}} \\
 & \textbf{train} & \textbf{test} & \textbf{train} & \textbf{test} & \textbf{train} & \textbf{test} & \textbf{train} & \textbf{test} & \textbf{train} & \textbf{test} & \textbf{train} & \textbf{test} & \textbf{train} & \textbf{test} \\
\midrule
Radar       & 2.0061  & 0.0204 & 1.8485  & 0.0213 & 1.9994  & 0.0229 & 5.8551  & 0.0865 & 0.4789  & 0.008  & 1.4281  & 0.1333 & 1.8316  & 0.4085 \\
ANOMALOUS   & 2.1543  & 0.0135 & 2.3067  & 0.0101 & 2.2285  & 0.0092 & 7.0728  & 0.027  & 1.252   & 0.0102 & 23      & 0.1734 & 70      & 0.5362 \\ \midrule
DOMINANT    & 2.3293  & 0.1633 & 2.6146  & 0.1886 & 2.8165  & 0.2138 & 11      & 0.8201 & 0.5937  & 0.0409 & 4.5385  & 0.4315 & 12      & 1.2824 \\
AnomalyDAE  & 113     & 1.1018 & 118     & 1.2149 & 134     & 1.4219 & 345     & 3.4737 & 32      & 0.3022 & 101     & 0.923  & 139     & 1.3636 \\
CoLA        & 247     & 20     & 196     & 18     & 172     & 15     & 1642    & 167    & 111     & 10     & 161     & 15     & 236     & 22     \\
CONAD       & 12      & 1.3023 & 13      & 1.3489 & 13      & 1.561  & 39      & 4.4816 & 1.85    & 0.586  & 14      & 1.0551 & 20      & 1.7071 \\
TAM         & 46      & 1.2557 & 27      & 1.4499 & 73      & 1.5347 & 44      & 5.4196 & 3.0339  & 0.2301 & 42      & 0.4958 & 59      & 0.8114 \\
PREM        & 0.4613  & 0.0066 & 0.3637  & 0.0075 & 0.3471  & 0.0065 & 0.508   & 0.0108 & 0.3056  & 0.0042 & 0.9174  & 0.0054 & 1.8403  & 0.0075 \\
GADAM       & 1.0099  & 0.0102 & 0.8122  & 0.0098 & 0.8889  & 0.0124 & 0.738   & 0.0213 & 0.7409  & 0.0054 & 1.6262  & 0.007  & 2.199   & 0.0085 \\
FIAD        & 227     & 2.0116 & 233     & 2.1419 & 242     & 2.1964 & 442     & 3.4145 & 67      & 0.497  & 215     & 1.9857 & 257     & 2.0871 \\
\midrule
\ourmethod & \multicolumn{2}{c|}{${\textbf{\underline{0.0317}}}$} & \multicolumn{2}{c|}{${\textbf{\underline{0.0338}}}$} & \multicolumn{2}{c|}{${\textbf{\underline{0.0141}}}$} & \multicolumn{2}{c|}{${\textbf{\underline{0.0357}}}$} & \multicolumn{2}{c|}{${\textbf{\underline{0.0601}}}$} & \multicolumn{2}{c|}{${\textbf{\underline{0.1641}}}$} & \multicolumn{2}{c}{${\textbf{\underline{0.3142}}}$} \\
\bottomrule
\end{tabular}%
}
\label{tab:training_testing_time}
\end{table*}

\begin{table*}[ht]
\caption{Maximum GPU memory usage (in MB) during training and testing for baselines and \ourmethod}
\centering
\resizebox{\textwidth}{!}{%
\begin{tabular}{l|cc|cc|cc|cc|cc|cc|cc}
\toprule
\multirow{2}{*}{\textbf{Method}} 
  & \multicolumn{2}{c|}{\textbf{Amazon}} 
  & \multicolumn{2}{c|}{\textbf{Reddit}} 
  & \multicolumn{2}{c|}{\textbf{Tolokers}} 
  & \multicolumn{2}{c|}{\textbf{YelpChi}} 
  & \multicolumn{2}{c|}{\textbf{Cora}} 
  & \multicolumn{2}{c|}{\textbf{BlogCatalog}} 
  & \multicolumn{2}{c}{\textbf{Flickr}} \\
 & \textbf{train} & \textbf{test} 
 & \textbf{train} & \textbf{test} 
 & \textbf{train} & \textbf{test} 
 & \textbf{train} & \textbf{test} 
 & \textbf{train} & \textbf{test} 
 & \textbf{train} & \textbf{test} 
 & \textbf{train} & \textbf{test} \\
\midrule
Radar       & 2814  & 2416  & 3257  & 2796  & 3715  & 3187  & 15202 & 13036 & 316   & 288   & 2248  & 1770  & 4797  & 3766  \\
ANOMALOUS   & 1223  & 1223  & 1426  & 1426  & 1605  & 1605  & 6548  & 6548  & 263   & 263   & 2426  & 2111  & 5185  & 4502  \\ \midrule
DOMINANT    & 2850  & 2035  & 3290  & 2351  & 3759  & 2682  & 15284 & 10910 & 272   & 228   & 1448  & 1225  & 3046  & 2567  \\
AnomalyDAE  & 3364  & 3335  & 3865  & 3844  & 4441  & 4403  & 17978 & 17945 & 343   & 338   & 1879  & 1847  & 3949  & 3903  \\
CoLA        &  423  &  404  &  487  &  467  &  552  &  532  &  2193 &  2173 &  79  &  63  &  432  &  376  &  922  &  729 \\
CONAD       & 2452  & 1639  & 2815  & 1876  & 3246  & 2173  & 13076   & 8706   &  253  &  167  & 1677  & 1185  & 3538  & 2488  \\
TAM         & 3968  & 3842  & 4536  & 4351  & 5150  & 4899  & 19742 & 17590 &  440  &  551  & 1535  & 1620  & 3036  & 3056  \\
PREM        &   91  &  34  &  109  &   43  &  114  &   49  &   219 &    84 &  111  &   59  & 1191  &   555  & 2510  & 1149  \\
GADAM       &   96  &   35  &   98  &  33  &  109  &   40  &   153 &    38  &  108  &   51  &  746  &   371  & 1515  &   756  \\
FIAD        & 4085  & 3281  & 4710  & 3784  & 5413  & 4351  & 21817   & 17471   &   450 &   364 & 2605  & 2073  & 5498  & 4361  \\
\midrule
\ourmethod
  & \multicolumn{2}{c|}{32}
  & \multicolumn{2}{c|}{30}
  & \multicolumn{2}{c|}{74}
  & \multicolumn{2}{c|}{30}
  & \multicolumn{2}{c|}{112}
  & \multicolumn{2}{c|}{1144}
  & \multicolumn{2}{c}{2445}
\\

\bottomrule
\end{tabular}%
}
\label{tab:training_testing_memory}
\end{table*}

\subsection{Experimental Setup}
\noindent\textbf{Datasets.} 
To validate the performance of our proposed method, we carried out comparative experiments on 10 benchmark datasets~\cite{tang2023gadbench,liu2021anomaly} across multiple domains, including 2 co-review networks (Amazon and YelpChi~\cite{mcauley2013amateurs,rayana2015collective}), 4 social networks (Reddit, Questions, BlogCatalog, and Flickr~\cite{kumar2019predicting,platonov2023critical,ding2019deep,tang2009relational}), a work collaboration network (Tolokers~\cite{platonov2023critical}), a transaction network (T-Finance~\cite{tang2022rethinking}), a payment flow network (Elliptic~\cite{weber2019anti}), and a citation network (Cora~\cite{sen2008collective}). Seven of them are datasets containing real-world anomalies, while the remaining three have artificially injected anomalies~\cite{ding2019deep}. The specific details of these datasets are outlined in Table~\ref{tab:dsets}. 

\noindent\textbf{Baselines.} We compare \ourmethod with both shallow methods and state-of-the-art (SOTA) deep methods for unsupervised graph anomaly detection (GAD). Specifically, we include 2 shallow GAD methods, i.e., Radar~\cite{li2017radar} and ANOMALOUS~\cite{peng2018anomalous}, as well as 8 deep GAD methods, i.e., DOMINANT~\cite{ding2019deep}, AnomalyDAE~\cite{fan2020anomalydae}, CoLA~\cite{liu2021anomaly}, CONAD~\cite{xu2022contrastive}, PREM~\cite{pan2023prem}, TAM~\cite{qiao2023truncated}, GADAM~\cite{chenboosting}, and FIAD~\cite{chen2025fiad}. Among them, the 2 shallow GAD methods are based on the PyGOD implementation~\cite{liu2024pygod,liu2022bond}, where all baselines hyperparameters are set based on the paper, and datasets not contained in the paper perform grid-specific random searches.

\noindent\textbf{Metrics.} 
Following the existing benchmark~\cite{tang2023gadbench}, we consider two metrics for evaluation: Area Under the Receiver Operating Characteristic Curve (AUROC) and Area Under the Prevision Recall Curve (AUPRC). We report the average AUROC/AUPRC across 5 trials.

\noindent\textbf{Implementation Details.} 
For hyper-parameter settings, we perform a specific set of random searches to select the key hyper-parameters in \ourmethod: $\alpha, \beta$: floats between $0$ and $1$, $L$: integers between $1$ and $20$, and $K$: integers between $10$ and $100$. 
The experiments were conducted using Python 3.8.19, CUDA 12.1, PyTorch 2.1.2, and DGL 1.1.3 on a Linux server with an Intel i5-13600K CPU and an NVIDIA RTX 3090 GPU (24GB). Our code is available at \url{https://github.com/yunf-zhao/FreeGAD}.

\subsection{Performance Comparison}
The comparison results on seven real-world graph datasets and three injected anomaly datasets (Cora, BlogCatalog, Flickr) are reported in Table~\ref{tab:combined}. From these results, we have the following observations. \ding{182}~\ourmethod demonstrates superior performance compared to SOTA approaches across seven real-world datasets, achieving better results on both evaluation metrics in most cases, with the sole exceptions being AUPRC on Elliptic and both AUROC and AUPRC on Reddit. These results demonstrate the strong effectiveness of our training-free method across various benchmarks.  \ding{183}~While \ourmethod does not achieve SOTA performance on the three injected anomaly datasets, it consistently outperforms traditional non-deep-learning models (e.g., Radar and ANOMALOUS) and maintains competitive performance with leading deep-learning approaches such as Dominant. Additionally, it's important to note that the community has recently deemed injection-based datasets as unreasonable \cite{huang2023unsupervised}. Existing methods might achieve better performance with specific designs, suggesting that future evaluations will likely focus on real anomalous datasets. \ding{184}~Due to its training-free nature, \ourmethod is not affected by random initialization or stochastic training processes, leading to its consistent and stable performance on all datasets. In contrast, training-based methods often suffer from performance fluctuations caused by randomness in training processes, such as weight initialization. \ding{185}~\ourmethod demonstrates strong scalability, achieving state-of-the-art results on large-scale datasets like T-Finance and Elliptic, where many baseline methods encounter out-of-memory (OOM) issues. This confirms \ourmethod's suitability for handling graphs with extensive data. 

\subsection{Efficiency Analysis}
To investigate the running efficiency of \ourmethod, we compare the whole runtime of \ourmethod with the training and testing time of baseline methods, and the results are illustrated in Table~\ref{tab:training_testing_time}. It is observed that \ourmethod eliminates the need for a training phase, which drastically reduces its overall computation cost. In contrast, training-based methods require significant time for model optimization, with some methods (e.g., CoLA and AnomalyDAE) taking hundreds of seconds on certain datasets, limiting their applicability. Moreover, \ourmethod achieves testing times comparable to, or even better than, many baseline methods. For instance, on BlogCatalog and Flickr datasets, its testing time is substantially lower than methods such as TAM and FIAD, highlighting its efficiency. 

In addition, we compare the peak memory usage during training and testing between \ourmethod and various baselines, as shown in Table~\ref{tab:training_testing_memory}. Notably, \ourmethod demonstrates significantly lower memory consumption on 4 out of 7 benchmark datasets. While CoLA, PREM, and GADAM come closest to \ourmethod, they still consume several times more memory on average, and other baselines often use tens to hundreds of times more. On the remaining three datasets (Cora, BlogCatalog, and Flickr), \ourmethod does not achieve the lowest memory usage, perhaps due to their high feature dimensionality (over 1,000 dimensions per node as shown in Table~\ref{tab:dsets}). Unlike methods that apply feature transformation or dimensionality reduction, \ourmethod operates directly on the raw high-dimensional data, resulting in higher memory consumption. Nonetheless, even on these challenging datasets, \ourmethod still outperforms most baselines, including DOMINANT and TAM. 

To sum up, \ourmethod achieves significant computational advantages without compromising on performance, highlighting its applications in scenarios where rapid deployment and adaptation are critical.

\subsection{Ablation Study}

\begin{table}[t]
\caption{Performance of \ourmethod and its variants in terms of AUROC (in percent). The best results are highlighted in bold and underlined.}
\centering
\begin{tabular}{l|cccc}
\midrule
\textbf{Variant} & \textbf{Cora} & \textbf{Reddit} & \textbf{YelpChi} & \textbf{Elliptic} \\
\midrule
\ourmethod & ${\textbf{\underline{84.85}}}$ & ${\textbf{\underline{57.21}}}$ & ${\textbf{\underline{78.55}}}$ & ${\textbf{\underline{77.25}}}$ \\
\midrule
w/o Multi-hop & 66.80 & 48.62 & 54.85 & 56.21 \\
w/o Anchor & 28.62 & 52.70 & 75.16 & 52.13 \\
w/o Negative & 30.83 & ${54.05}$ & 78.55 & 52.84 \\
w/o Positive & ${84.83}$ & 48.24 & 25.58 & ${65.61}$ \\
\midrule
w/ Max Score & 70.21 & 54.03 & 73.85 & 76.48 \\
w/ Min Score & 77.34 & 56.78 & 75.34 & 76.34 \\
w/ Avg Score & 84.68 & 55.99 & 77.25 & 41.37 \\
\bottomrule
\end{tabular}
\label{tab:ablation_v1}
\end{table}

To verify the effectiveness of each key component and design in \ourmethod, we conduct an ablation study by introducing several variants of \ourmethod for comparison.

We compare \ourmethod with four variants that each exclude a key component: \ding{182}~\textbf{w/o Multi-hop}, where the representations at the last layer serve as the final representations; \ding{183}~\textbf{w/o Anchor}, where anchor nodes are randomly selected rather than using affinity-based selection; \ding{184}~\textbf{w/o Negative} and \textbf{w/o Positive}, where only positive or negative score contributes to the final anomaly score. In the upper part of Table~\ref{tab:ablation_v1}, we can observe that all key components of \ourmethod contribute significantly to its performance. Removing the multi-hop mechanism (\textbf{w/o Multi-hop}) results in a substantial performance drop across all datasets, confirming the effectiveness of our affinity-gated residual encoder designed under TAM’s ``one-class homophily'' principle~\cite{qiao2023truncated}. Specifically, the encoder leverages this principle to amplify the differences between normal and anomalous nodes, facilitating anomaly pattern detection. The performance degradation in the \textbf{w/o Anchor} variant demonstrates that affinity-based anchor selection is crucial for effectively distinguishing anomalies. Furthermore, excluding either positive or negative scoring (\textbf{w/o Positive} or \textbf{w/o Negative}) leads to noticeable declines, indicating the complementary role of both scores in achieving robust anomaly detection.

Furthermore, we make separated evaluation using only the maximum score (\textbf{w/ Max Score}), only the minimum score (\textbf{w/ Min Score}), and only the average score (\textbf{w/ Avg Score}) to demonstrate the benefit of their combination (lower part of Table~\ref{tab:ablation_v1}). On the citation network (Cora), the Avg outperforms both Max and Min, more effectively capturing anomalous behaviors. In social networks (Reddit), Min excels at detecting extreme social anomalies. On the co-review network (YelpChi), Avg again leads by more accurately identifying anomalous reviews. For fraud detection (Elliptic), Max and Min both surpass Avg by pinpointing subtle deviations in financial behavior. Overall, combining all three metrics yields a more comprehensive anomaly score that consistently outperforms any single measure across diverse domains.

\begin{figure} [tbp]
    \centering
    \subfigure[Cora\label{subfig:hp_cora}]{
        \includegraphics[scale=0.18]{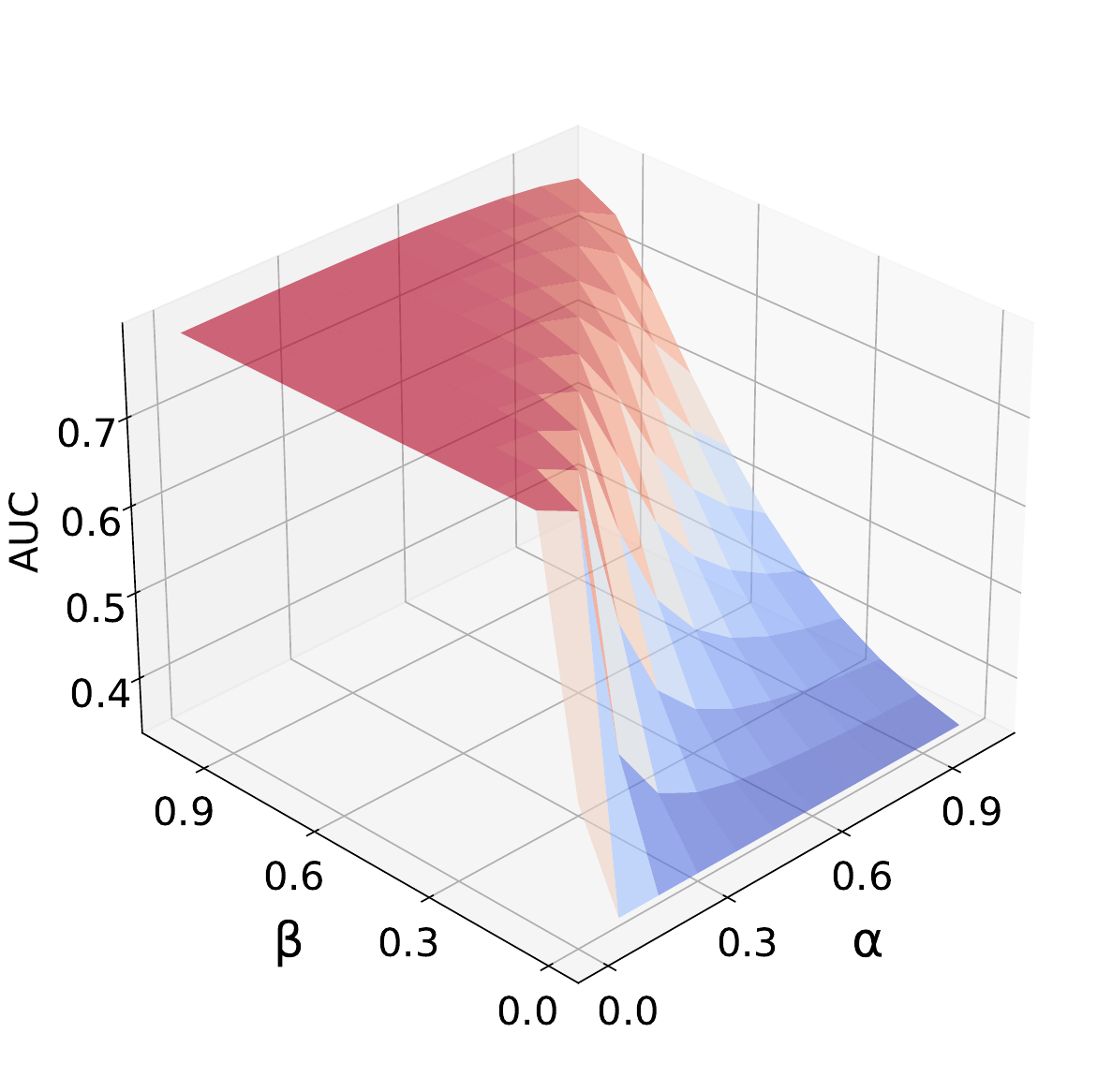}
    }\hfill
    \subfigure[Questions\label{subfig:hp_questions}]{
        \includegraphics[scale=0.18]{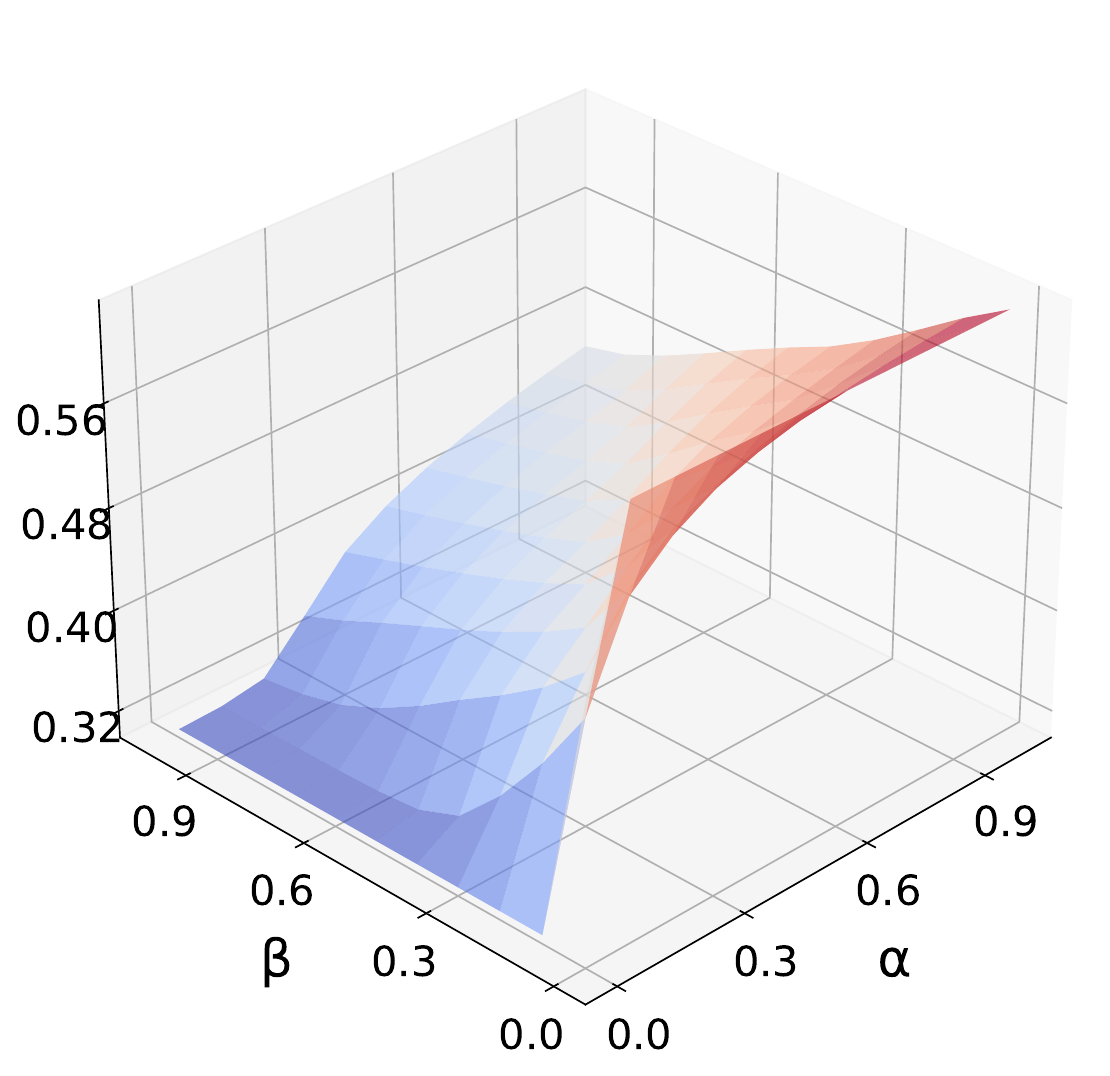}
    }\hfill
    
    \subfigure[YelpChi\label{subfig:hp_yelpchi}]{
        \includegraphics[scale=0.18]{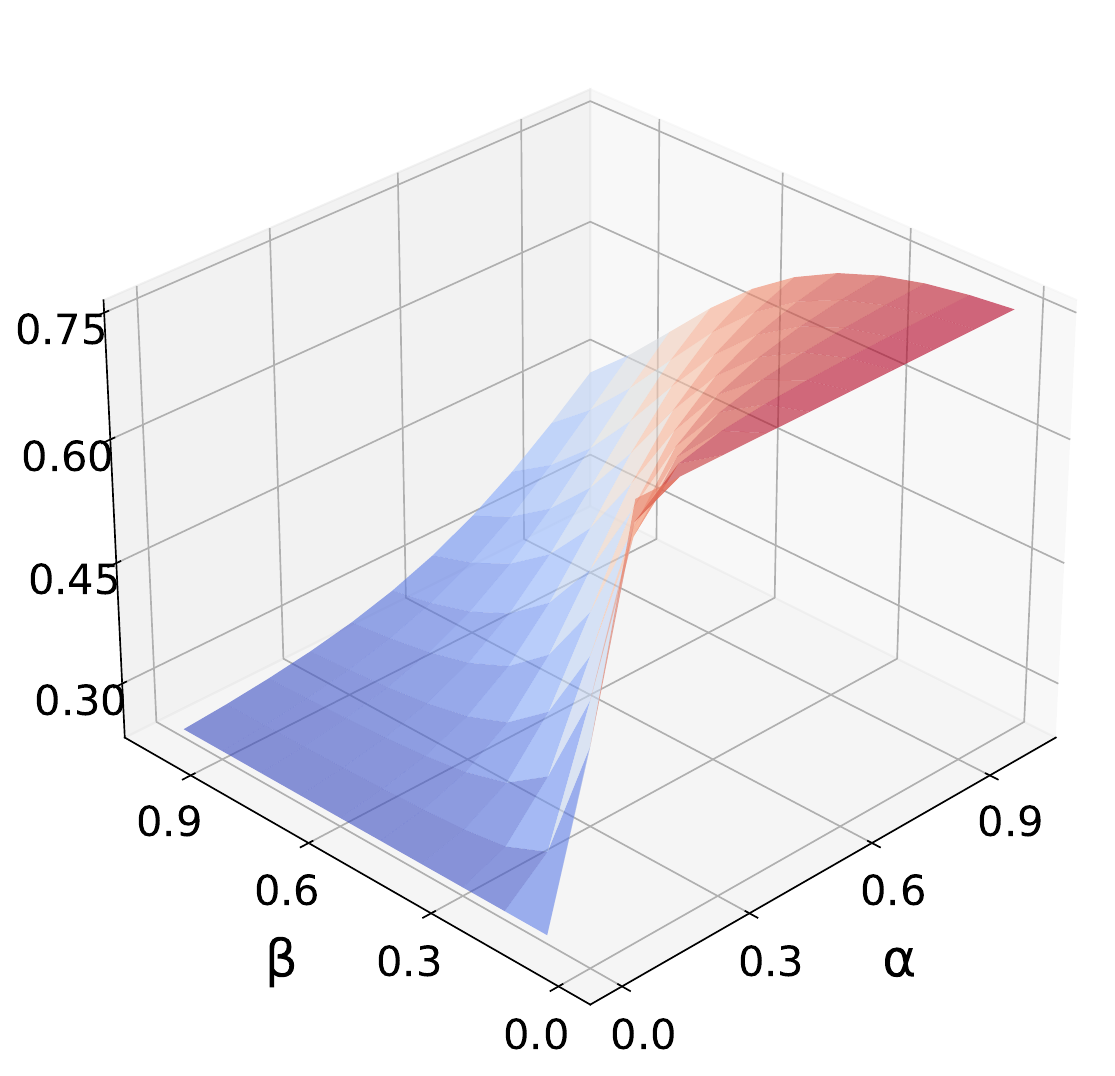}
    }\hfill
    \subfigure[Reddit\label{subfig:hp_reddit}]{
        \includegraphics[scale=0.18]{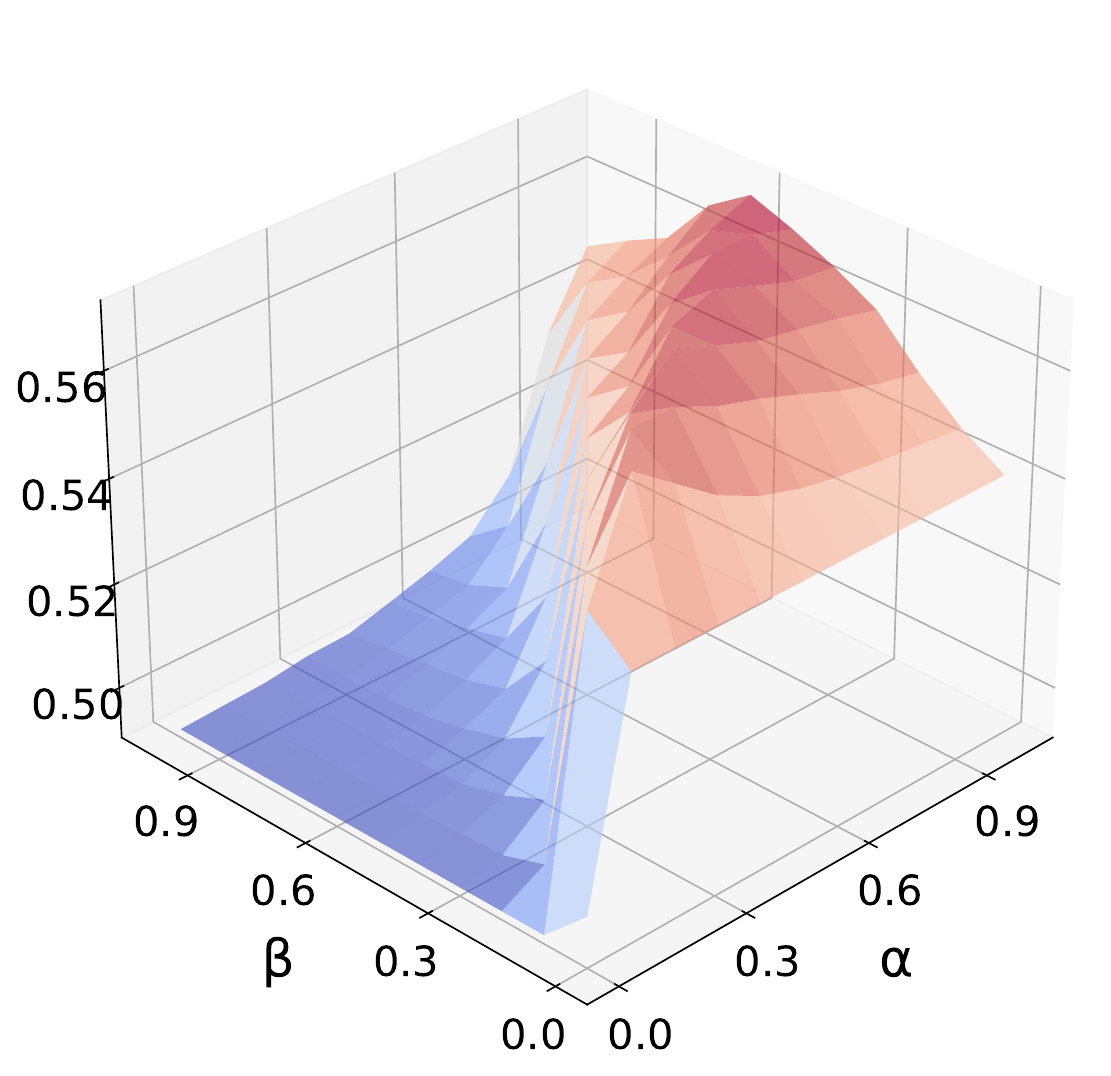}
    }\hfill
    \caption{The sensitivity of \ourmethod in terms of $\alpha$ and $\beta$.}
    \label{fig:hyperparams}
\end{figure}

\begin{figure} [t]
    \centering
    \subfigure[Cora\label{subfig:hop_cora}]{
        \includegraphics[scale=0.35]{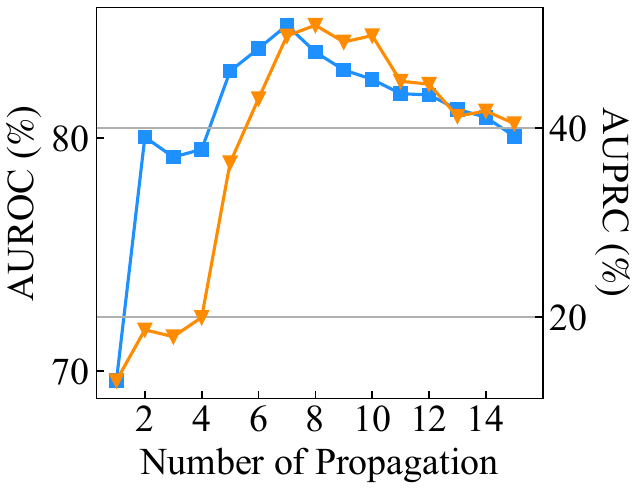}
    }\hfill
    \subfigure[Questions\label{subfig:hop_questions}]{
        \includegraphics[scale=0.35]{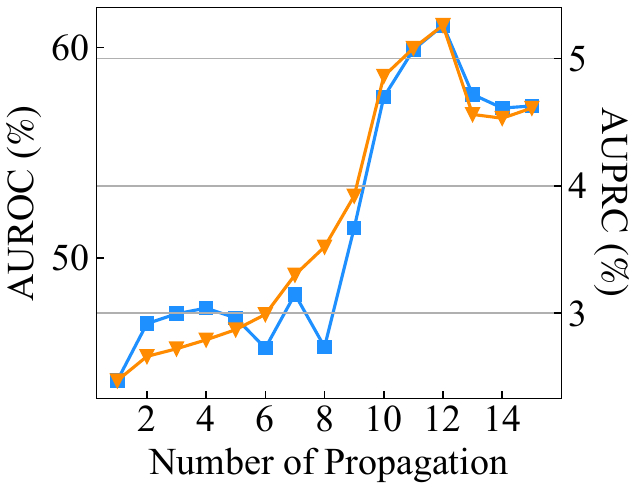}
    }
    \subfigure[YelpChi\label{subfig:hop_YelpChi}]{
        \includegraphics[scale=0.35]{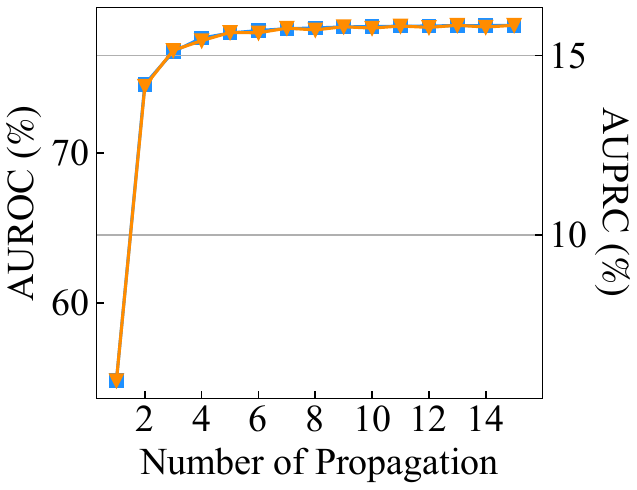}
    }\hfill
    \subfigure[Reddit\label{subfig:hop_Reddit}]{
        \includegraphics[scale=0.37]{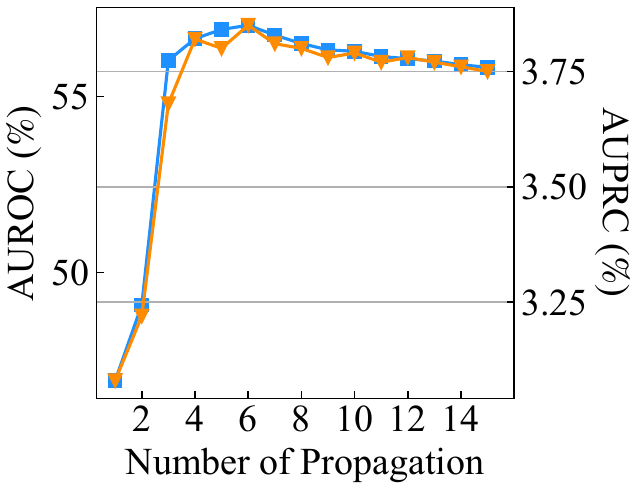}
    }
    \caption{The sensitivity of \ourmethod on layer number $L$.}
    \label{fig:prop}
\end{figure}
\subsection{Parameter Analysis}
We investigate the impact of key hyper-parameters, i.e., the scoring trade-off weight $\alpha$ and $\beta$ (in Eq.~\eqref{eq:score}), the propagation layer $L$ and the anchor nodes number $K$ to the performance of \ourmethod. 

\noindent\textbf{Trade-off Weight $\alpha$ and $\beta$.} The results of sensitivity study for $\alpha$ and $\beta$ are shown in Fig.~\ref{fig:hyperparams}. According to the results, we can see that for different datasets, the reliances on $\alpha$ and $\beta$ are quite different. For example, Cora(citation) requires a higher $\beta$, indicating the contribution of negative scores; in contrast, Questions(social), Reddit(social) and YelpChi(co-review) requires a higher $\alpha$, highlighting the importance of positive scores in anomaly detection. This disparity suggests that the optimal balance between positive and negative contributions varies depending on the dataset's domain-specific anomaly patterns.

\noindent\textbf{Propagation Layer $L$.} Fig.~\ref{fig:prop} shows the effect by the propagation layer $L$, where a larger $L$ (usually $L\geq8$) is preferred for all four datasets.
This indicates that considering multi-hop neighbors is crucial for effective anomaly detection in \ourmethod. Meanwhile, it also indicates that the shallow GNNs in existing GAD methods may lead to suboptimal performance. 

\section{Conclusion}
In this paper, we proposed \ourmethod, a novel training-free method for graph anomaly detection (GAD) that eliminates the need for resource-intensive training processes, addressing the scalability and deployment challenges of traditional approaches. \ourmethod uses an affinity-gated residual encoder to generate anomaly-aware representations and identifies anchor nodes as pseudo-normal and anomalous guides for effective anomaly scoring. Extensive experiments on benchmark datasets demonstrate that \ourmethod achieves superior performance, efficiency, and scalability compared to state-of-the-art methods, which highlights the potential of training-free approaches in advancing GAD research.

\begin{acks}
This work was partially supported by the Specific Research Project of Guangxi for Research Bases and Talents (GuiKe AD24010011), the Key Research \& Development Program Project of Guangxi (GuiKe AB25069095).
\end{acks}

\section*{GenAI Usage Disclosure}
Generative AI tools were utilized solely for the purpose of language polishing in this paper. All other components of the research, including but not limited to code, figures, and tables, were developed without the assistance of Generative AI tools.

\appendix

\bibliographystyle{ACM-Reference-Format}
\balance
\bibliography{arxiv}

\end{document}